\documentclass[sigconf]{acmart}

\AtBeginDocument{%
  \providecommand\BibTeX{{%
    \normalfont B\kern-0.5em{\scshape i\kern-0.25em b}\kern-0.8em\TeX}}}

\setcopyright{rightsretained}
\copyrightyear{2024}
\acmMonth{December}
\acmYear{2024}

\begin{document}

\title{Agents Are Not Enough}

\author{Chirag Shah}
\orcid{0000-0002-3797-4293}
\email{chirags@uw.edu}
\affiliation{%
  \institution{University of Washington, Seattle, WA, USA}
}

\author{Ryen W. White}
\orcid{0000-0002-3797-4293}
\email{ryenw@microsoft.com}
\affiliation{%
  \institution{Microsoft Research, Redmond, WA}
  \country{USA}
}

\renewcommand{\shortauthors}{Shah \& White}

\begin{abstract}
In the midst of the growing integration of Artificial Intelligence (AI) into various aspects of our lives, agents are experiencing a resurgence. These autonomous programs that act on behalf of humans are neither new nor exclusive to the mainstream AI movement. By exploring past incarnations of agents, we can understand what has been done previously, what worked, and more importantly, what did not pan out and why. This understanding lets us to examine what distinguishes the current focus on agents. While generative AI is appealing, this technology alone is insufficient to make new generations of agents more successful. To make the current wave of agents effective and sustainable, we envision an ecosystem that includes not only agents but also Sims, which represent user preferences and behaviors, as well as Assistants, which directly interact with the user and coordinate the execution of user tasks with the help of the agents.
\end{abstract}



\maketitle

\section{Introduction}
An agent, in the context of AI, is an autonomous entity or program that takes preferences, instructions, or other forms of inputs from a user to accomplish specific tasks on their behalf. Agents can range from simple systems, such as thermostats that adjust ambient temperature based on sensor readings, to complex systems, such as autonomous vehicles navigating through traffic. Key characteristics of agents include autonomy, programmability, reactivity, and proactiveness. The area of agentic AI covers AI systems designed to operate with a high degree of autonomy, making decisions and taking actions independently of human intervention.

Agents, by definition, remove agency from a user in order to do things on the user's behalf and save them time and effort. However, such a trade-off may be brought into question if relinquishing that control does not generate sufficient user value. The agents may also make mistakes, require intervention or supervision, and may be limited to performing only simple tasks. These shortcomings are evident from the agentic research and development efforts over the past decade or so. Agents, often referred to as operators, skills, apps, extensions, and plugins, have been widely available through integrations into computers, smartphones, speakers, wearables, and  automobiles. However, their utility has been severely limited \cite{kurz2021success, alagha2019evaluating}. In addition to the limited applications, there are continuing shortcomings that these agents exhibit that are not addressed by simply creating more capable systems. Here, we briefly review why this is the case and what we can do about it. Specifically, we argue that while making agents more capable will address some of the issues, it will not be enough. We need to build a whole new ecosystem with highly capable agents that could tackle complex tasks on a user's behalf, while ensuring privacy and trustworthiness.

\section{Historical Attempts and Failures}
There are five distinct eras of agents development we can identify, each differentiated by the core architecture or technology being used and the challenges for widespread success.

\noindent
{\bf Early AI Agents}\\
The idea of AI agents dates back to the 1950s with symbolic AI. Early examples, such as the General Problem Solver (GPS), aimed to replicate human problem-solving using symbolic reasoning. However, these agents struggled with real-world complexity due to their dependence on predefined rules and lack of adaptability \cite{huang2024levels}.

\noindent
{\bf Expert Systems}\\
In the 1980s, expert systems like MYCIN and DENDRAL emerged, utilizing domain-specific knowledge for decision-making. While effective in narrow domains, these systems were brittle and unable to generalize beyond their programmed expertise. The extensive manual knowledge engineering required made them impractical for broader applications \cite{kapoor2024ai}.

\noindent
{\bf Reactive Agents}\\
The 1990s introduced reactive agents \cite{masterman2024landscape}, which responded to environmental stimuli without internal models. Rodney Brooks’ subsumption architecture \cite{brooks1986robust} exemplified this approach, emphasizing real-time interaction over complex reasoning. However, reactive agents lacked the ability to plan or learn from past experiences, limiting their utility in dynamic environments.

\noindent
{\bf Multi-Agent Systems}\\
Multi-agent systems (MAS) \cite{xie2017multi} brought the concept of multiple interacting agents, each with specific roles. While MAS showed promise in distributed problem-solving, they faced challenges in coordination, communication, and scalability. Managing interactions among agents often led to inefficiencies and unpredictable behaviors.

\noindent
{\bf Cognitive Architectures}\\
Cognitive architectures like SOAR and ACT-R \cite{laird2022analysis} aimed to model human cognition, integrating perception, memory, and reasoning. Despite their sophisticated designs, these architectures struggled with scalability and real-time performance. Their complexity often resulted in high computational costs and limited practical applications.

Not all of these efforts have panned out. Some of the simple agents that use rule-based systems and symbolic logic are widely used with limited capabilities (e.g., Alexa, Siri), and some of the multi-agent frameworks such as Swarm Robotics \cite{brambilla2013swarm} and AutoGen \cite{wuautogen} have had various successes in solving complex tasks within limited domains. However, we lack agentic systems that could score high on capabilities (e.g., solving complex tasks) as well as applicability (wide range of scenarios, modalities, and contexts). We believe this is for the following five reasons.

\begin{enumerate}
    \item {\bf Lack of generalization}. Many AI agents are designed for specific tasks and fail to generalize across different domains. This limitation arises from their reliance on predefined rules and lack of adaptive learning mechanisms.
    \item {\bf Scalability issues}. As the complexity of tasks increases, the computational resources required by AI agents grow exponentially. This scalability issue hampers their ability to handle real-world applications effectively.
    \item {\bf Coordination and communication}. In multi-agent systems, effective coordination and communication are critical. However, ensuring seamless interaction among agents remains a significant challenge, often leading to inefficiencies and conflicts. In addition, we also need enhanced mechanisms between a user and an agent to ensure that the questions and recommendations provided by the agent are both appropriate and effective for the task at hand.
    \item {\bf Robustness}. Many AI agents are brittle, meaning they perform well under specific conditions but fail when faced with unexpected situations. This brittleness stems from their lack of robust learning and adaptation capabilities.
    \item {\bf Ethical concerns and safety}. Ensuring that AI agents operate ethically and safely is a major concern. Failures in this area can lead to unintended consequences, such as biased decision-making or harmful actions. In addition, the trade-offs resulting from giving an agent more control to accomplish a task at the expense of user agency and learning opportunities are not well understood. 
\end{enumerate}

\section{Can we fix agents?}
Before dismissing agents or agentic AI as a passing trend, it is important to consider how their shortcomings can be effectively addressed. That may not be enough, but it is a start. So what can we do? We propose five directions that roughly correspond to the five failures listed above.

\begin{enumerate}
    \item {\bf Integrating machine learning and symbolic AI}. Combining machine learning with symbolic AI can enhance the adaptability and reasoning capabilities of AI agents. Machine learning can provide the flexibility to learn from data, while symbolic AI can offer structured reasoning and explainability.
    \item {\bf New architectures}. Implementing caching solutions that store and execute agent workflows and reduce the need for calls to foundation models for common tasks, and new hybrid and hierarchical architectures that integrate small language models and large language models, can improve scalability and efficiency. By decomposing tasks into sub-tasks and assigning them to specialized agents, we can manage complexity more effectively.
    \item {\bf Enhanced coordination mechanisms}. Developing advanced coordination mechanisms, such as decentralized control and negotiation protocols, can improve the performance of multi-agent systems. These mechanisms can facilitate better communication and collaboration among agents.
    \item {\bf Robust learning algorithms}. Incorporating robust learning algorithms, such as reinforcement learning and transfer learning, can enhance the adaptability of AI agents. These algorithms enable agents to learn from their experiences and apply knowledge across different tasks.
    \item {\bf Ethical and responsible design}. Ensuring ethical and safe AI design involves implementing guidelines and frameworks that prioritize transparency, fairness, and accountability. Integrating explainability into system design and robust testing with system deployment can help mitigate ethical and safety concerns.
\end{enumerate}

\section{Why agents are not enough}
The previous section may create an impression that if only we could address those technical challenges, we could finally have life-changing agents. While those challenges are not trivial to meet, it will not be sufficient to enable the rise of capable and widespread agents. There are more issues to address beyond those covered by technologies alone. Specifically, for agents to be successful, we will need to pay attention to at least the following five aspects.

\begin{enumerate}
    \item {\bf Value generation}. An agent is meant to provide autonomous execution of tasks on a user's behalf, but there are costs and risks. For instance, if the user needs to intervene or clarify frequently, that may defeat the purpose of an agent. The user may also face the trade-off between privacy and utility regarding the agent. In short, without the user realizing enough value out of an agent, they may not be willing to use it. Here, value can be understood as the difference between the perceived benefit and the perceived cost (e.g., time, privacy) of using an agent. 
    \item {\bf Adaptable personalization}. Every user and every situation is different when it comes to executing the task. An agent that cannot adapt to the user or their context may be of limited use. For instance, what if performing an online transaction on a user's behalf calls for resetting a password? The agent will need to be capable of doing this, but more importantly, depending on the task, the situation, and knowledge about the user, the agent could proceed with this subtask on its own or seek the user's input.
    \item {\bf Trustworthiness}. The more capable an agent is, the more the user will need to be able to trust it. Letting agents perform bank transactions, personal communications, and important decision-making tasks will call for stronger scrutiny of and well-placed trust in those agents. This trust will also not be built overnight. Rather, through increased accuracy and transparency, the agents will have to gradually earn our trust. We still do not have a broad acceptance of automatically generated emails. Having AI-based agents perform more than content generation will require much more familiarity with and trust in those systems.
    \item {\bf Social acceptability}. We envision a future where agents can do many tasks on a user's behalf, including shopping, scheduling, and negotiating. However, to have these done at scale and for diverse populations, cultures, and customs, we need to have wide social acceptability of agent-based interactions and transactions. This may take a long time to materialize. For instance, while paying bills online offers many advantages to individuals, service providers, and the environment and many in the developed world are accustomed to using it, there is still a significant fraction of the world where this is not a common practice for various reasons.
    \item {\bf Standardization}. Developing and deploying agents is and will continue to be decentralized, which is desired for a sustained ecosystem around agents. However, this will also pose new challenges regarding compatibility, reliability, and security of those agents. Therefore, we will need efforts to standardize how agents are deployed, connected (in case of multi-agent frameworks), and served. Consider this similar to developing a networking protocol or an app store.
\end{enumerate}

\section{A new ecosystem with agents}
To overcome the challenges of agentic AI listed above, we need three specific mechanisms:

\begin{enumerate}
    \item A private and secure version of an agent to ensure user information is protected while making both private and public versions of agents tackle more meaningful and complex tasks.
    \item A representation of user that can interact with an agent on a user's behalf so the user does not have to keep intervening or providing frequent inputs.
    \item Ability for an agent with intimate knowledge of users and their tasks to communicate and negotiate with other agents on a user's behalf to accomplish complex tasks without added burden on the user.
\end{enumerate}

\begin{figure}[htbp]
  \centering
  \includegraphics[width=1.0\linewidth]{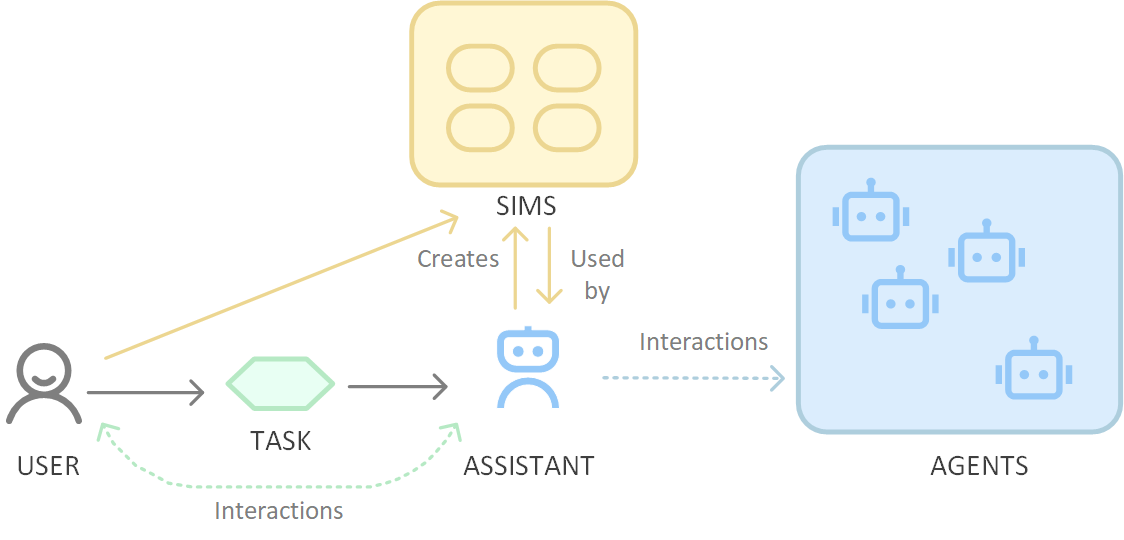}
  \caption{Envisioning a new eco-system with Agents, Sims, and Assistants.}
  \label{fig:eco-system}
\end{figure}

To put these recommendations in practice, we propose a new kind of ecosystem that is built around agents, but also includes other critical components that provide standardization, privacy, personalization, and increased trust. Specifically, we envision an ecosystem depicted in Figure \ref{fig:eco-system} that comprises of Agents, Sims, and Assistants.

Agents are narrow and purpose-driven modules that are trained to do a specific task. Each agent can be autonomous, but with an ability to interface with other agents. Section 3 provides suggestions for how to improve agents. 

Sims are representations of a user. Each Sim is created using a combination of user profile, preferences, and behaviors, and captures an aspect of who the user is. Different Sims can have different privacy and personalization settings. In that regard, they act more than a user persona (which represents a target audience) or profile (which usually focuses on topical or domain interests); they carry awareness of the user's preferences, privacy, and the contexts in which the user operates. A Sim can also act, that is, a Sim can interact with an agent on user’s behalf to accomplish a task. This is coordinated by the user’s Assistant.

An Assistant is a program that directly interacts with the user, has a deep understanding of that user, and has an ability to call Sims and Agents as needed to reactively or proactively accomplish tasks and sub-tasks for the user. In this regard, an Assistant is a private version of an agent that can have access to a user's personal information and could be fine-tuned to that user, allowing it to act on the user's behalf.

The interaction between Agents, Sims, and Assistants is characterized by a high degree of synergy. The Assistant, with its comprehensive understanding of the user, co-creates and manages Sims with the supervision of the user, reflect the user’s multifaceted life. These Sims, in turn, engage with specialized Agents to perform tasks efficiently. This layered approach ensures that tasks are handled with precision and personalization, enhancing overall user satisfaction.

\section{Future of Agentic AI}
We believe agents represent the next age of evolution for capable AI systems. However, as we have argued, simply building capable agents is not going to ensure their wide applicability and acceptability. On one hand, we need to build more capable agents that go beyond information retrieval and generation to doing reasoning and taking actions on a user's behalf. And on the other hand, we need to develop various mechanisms to enable more meaningful interactions among agents as well as a user and agents (what we refer to as Assistants here). We also need to address personalization, privacy, user agency, value generation, and trustworthiness of agents.

To address these issues, we envision a new ecosystem that includes different kind of agents, but more importantly, constructs such as Sims and Assistants. We believe similar to an app store, there could be an agent store with vetted agents available for a user or their Assistants to interact with and accomplish various tasks. Agents may be the centerpiece of this ecosystem, but it is the availability of Sims, Assistants, and the set of protocols that connects them that can really make this next evolution of agentic AI successful.

\bibliographystyle{ACM-Reference-Format}

\end{document}